\documentclass[conference]{IEEEtran}
\IEEEoverridecommandlockouts
\usepackage{cite}
\usepackage{amsmath,amssymb,amsfonts}
\usepackage{algorithmic}
\usepackage{graphicx}
\usepackage{textcomp}
\usepackage{xcolor}
\usepackage{url} 
\usepackage{enumitem}

\def\BibTeX{{\rm B\kern-.05em{\sc i\kern-.025em b}\kern-.08em
    T\kern-.1667em\lower.7ex\hbox{E}\kern-.125emX}}

\begin{document}

\title{InteractiveGNNExplainer: A Visual Analytics Framework for Multi-Faceted Understanding and Probing of Graph Neural Network Predictions
}

\author{
\IEEEauthorblockN{TC Singh}
\IEEEauthorblockA{\textit{Department of Electrical Engineering} \\
\textit{Indian Institute of Technology}\\
Delhi, India \\
tcsingh24@gmail.com}
\and
\IEEEauthorblockN{Sougata Mukherjea}
\IEEEauthorblockA{\textit{Department of Electrical Engineering} \\
\textit{Indian Institute of Technology}\\
Delhi, India \\
sougatam@iitd.ac.in}
}

\maketitle

\begin{abstract}
Graph Neural Networks (GNNs) excel in graph-based learning tasks, but their complex, non-linear operations often render them as opaque ``black boxes''. This opacity hinders user trust, complicates debugging, bias detection, and adoption in critical domains requiring explainability. This paper introduces InteractiveGNNExplainer, a visual analytics framework to enhance GNN explainability, focusing on node classification. Our system uniquely integrates coordinated interactive views (dynamic graph layouts, embedding projections, feature inspection, neighborhood analysis) with established post-hoc (GNNExplainer) and intrinsic (GAT attention) explanation techniques. Crucially, it incorporates interactive graph editing, allowing users to perform a ``what-if'' analysis by perturbing graph structures and observing immediate impacts on GNN predictions and explanations. We detail the system architecture and, through case studies on Cora and CiteSeer datasets, demonstrate how InteractiveGNNExplainer facilitates in-depth misclassification diagnosis, comparative analysis of GCN versus GAT behaviors, and rigorous probing of model sensitivity. These capabilities foster a deeper, multifaceted understanding of GNN predictions, contributing to more transparent, trustworthy, and robust graph analysis.
\end{abstract}


\section{Introduction}
Graph Neural Networks (GNNs) \cite{zhou2020} have become pivotal for the analysis of graph-structured data, achieving state-of-the-art results in diverse fields such as social network analysis \cite{ham2017} and bio-informatics \cite{gil2017}. By aggregating neighborhood information, GNNs capture intricate features and topological patterns. However, their multilayer, non-linear message passing often obscures their decision-making processes, limiting their adoption where transparency is crucial for debugging, trust, fairness, and scientific discovery. The challenge lies in bridging the gap between the GNN's powerful predictive capabilities and the human user's need to understand and verify its reasoning.

Existing eXplainable AI (XAI) methods for GNNs, such as GNNExplainer \cite{ying2019} or intrinsic GAT attention\cite{vel2018}, provide valuable but often static and isolated insights. Interpreting these numerical or structural outputs effectively requires contextualization and interactive exploration. 

Visual analytics (VA) \cite{keim2008} offers a powerful solution by merging automated analysis with interactive visualizations. A key aspect, often underexplored in the GNN XAI tools, is allowing users to perform ``what-if'' analyses by directly manipulating graphs to understand the sensitivity of the model and the causal underpinnings. This active user involvement is critical for building robust mental models of GNN behavior.

This paper introduces \textbf{InteractiveGNNExplainer}, a Visual Analytics framework for multifaceted, interactive understanding and probing of GNN node classification predictions. We address these research questions:
\begin{itemize}
    \item \textbf{RQ1 (Enhanced Comprehension)}: How can synergizing diverse views (algorithmic explanations, graph structure, embeddings, features, neighborhood) yield a more comprehensive GNN prediction understanding than standalone methods?
    \item \textbf{RQ2 (Effective Model Probing)}: How can interactive graph editing with re-inference serve as an effective mechanism for probing model sensitivity and validating explanation-derived hypotheses?
    \item \textbf{RQ3 (Diagnosis and Comparison)}: Can such a framework facilitate the diagnosis of GNN misclassification and comparative analysis of different GNN architectures (GCN vs. GAT)?
\end{itemize}
InteractiveGNNExplainer integrates dynamic graph visualization, interactive embedding exploration, feature inspection, neighborhood analysis, GNNExplainer/GAT attention displays, and graph editing with immediate re-inference. Our contributions are as follows. 
\begin{itemize}
\item A novel Visual Analytics framework combining multiple explanation modalities with interactive graph perturbation. 
\item A detailed system architecture leveraging PyTorch Geometric~\cite{fey2019}. 
\item Case studies demonstrating the diagnosis of misclassifications, GCN/GAT comparison, and model sensitivity probing.
\end {itemize}

The rest of the paper is organized as follows. The next section cites related work. Section \ref{sec:framework} describes the InteractiveGNNExplainer system. Section \ref{sec:evaluation} evaluates the system using sone case studies. Section \ref{sec:discussion} has a discussion of the research questions that we are addressing in this paper. Finally, Section \ref{sec:conclusion} concludes the paper.

\section{Related Work}
\label{sec:related}
Our work is related to previous research in GNN explanation methods and Visual Analytics systems for deep learning. Our focus is on an instance-level explanation, with the aim of clarifying individual model decisions.

\subsection{GNN Explanation Methods}
GNNExplanations \cite{yuan2023} make predictions understandable.

\subsubsection{Perturbation-based Methods}
These methods, like GNNExplainer \cite{ying2019}, assess the importance of the input component by observing changes in the output of the model upon their perturbation. GNNExplainer learns sparse masks for edges and node features within a node's computational subgraph, optimizing for mutual information with the original prediction while encouraging sparsity. Our system directly integrates and visualizes its outputs. Other methods include ZORRO \cite{funke2023} and SubgraphX \cite{yuan2021}. Although powerful, their sensitivity to search space and computational cost for large neighborhoods are limitations.

\subsubsection{Gradient-based and Feature Attribution Methods}
Inspired by computer vision, these methods use output gradients with respect to inputs. Examples include IntegratedGradients \cite{sun2017} and graph adaptations such as GraphLIME \cite{huang2023}, which builds interpretable local surrogates. These methods can produce diffuse-saliency maps on graphs.

\subsubsection{Surrogate Models}
These approximate a complex GNN with a simpler, interpretable model (e.g., decision tree) \cite{yuan2023}. PGM-Explainer~\cite{vu2020} uses probabilistic graphical models. The fidelity of the surrogate is key to the explanation's validity.

\subsubsection{Attention-based Explanations}
GNNs with attention, such as GAT \cite{vel2018}, provide attention coefficients as intrinsic explanations for neighbor importance. InteractiveGNNExplainer visualizes these, offering a model-native perspective.

\subsubsection{Counterfactual and Factual Reasoning}
These methods find minimal input changes altering a prediction (counterfactual) \cite{lucic2022} or minimal evidence for the current prediction (factual). Our graph editing allows for manual exploration of similar ``what-if" scenarios.

These methods, while diverse, often produce outputs that require sophisticated interactive tools for effective user interpretation and hypothesis generation.

\subsection{Visual Analytics for Deep Learning}
Visual Analytics \cite{keim2008} aids sense-making in complex data. Several visual analysis systems have been developed for Deep Learning \cite{hoh2019}. For GNNs, GNNLens \cite{jin2023} diagnoses errors via multi-level views. CorGIE \cite{liu2022} visualizes the graph-to-embedding correspondence. InteractiveGNNExplainer extends these systems by uniquely combining GNNExplainer and GAT attention with interactive graph editing. This emphasis on direct perturbation and immediate feedback for causal probing and model comparison (GCN vs GAT) within a unified framework is a key differentiator for our system, enabling a more active human-in-the-loop analytical process.

\section{InteractiveGNNExplainer System}
\label{sec:framework}
InteractiveGNNExplainer is a Python Dash web application, using Plotly for charting and Dash Cytoscape for graph visualization. The backend leverages PyTorch and PyTorch Geometric~\cite{fey2019} for GNN operations and explanations, particularly the \texttt{torch\_geometric.explain} module.

\subsection{Design Goals and Principles}
The system prioritizes: 
\begin{itemize}
    \item Multi-Faceted Insights: Offering holistic understanding through complementary structural, feature-based, embedding, and algorithmic views. 
    \item Interactivity \& Coordination: Enabling dynamic exploration with seamlessly linked views where user selections propagate. 
    \item Direct Model Probing: Allowing "what-if" analyses via graph editing with immediate feedback on predictions and explanations. 
    \item Comparative analysis: Comparison of GCN and GAT models and their explanations. 
    \item Simplicity: Providing an intuitive interface for users with varying XAI expertise, aiming to lower the barrier to entry for complex GNN analysis.
\end{itemize}

\subsection{System Architecture}
\begin{figure}[htbp]
\centering
\includegraphics[width=0.8\columnwidth]{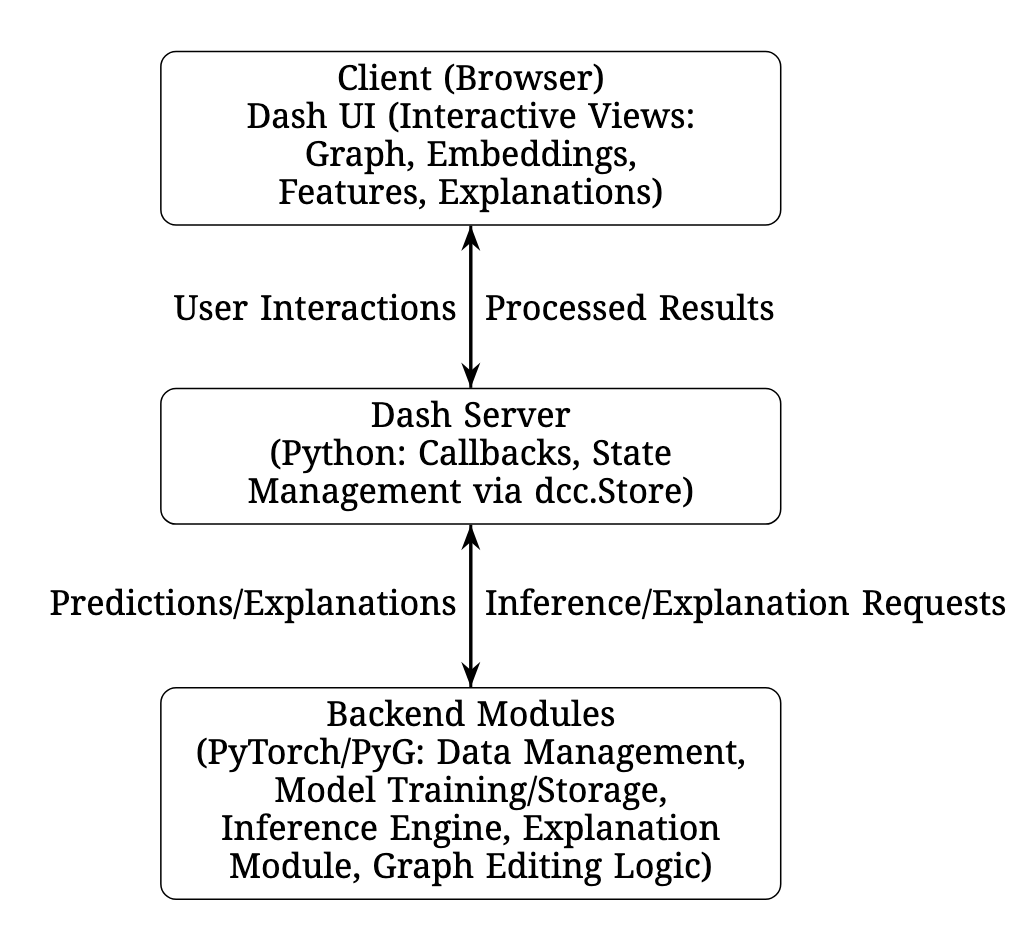} 
\caption{High-level architecture: User interacts with UI; Dash Callbacks trigger Backend Modules (Data, Model, Inference, Explanation, Editing) which update UI views.}
\label{fig:architecture}
\end{figure}

Fig.~\ref{fig:architecture} shows the client-server architecture of the stsyem. The client (browser) handles UI rendering and user interactions, while the Python Dash server manages data, model computations, and explanation logic.

\subsubsection{Backend (Python)}
\begin{itemize}
    \item \textit{Data Management}: The system loads PyG data sets (Cora, CiteSeer, AmazonPhoto), prioritizing preprocessed \texttt{.pt} files for speed. It then manages graph data (\texttt{x, edge\_index, y, masks}) and metadata for Dash app via \texttt{dcc.Store}.
    \item \textit{Model Training \& Storage}: We perform offline pre-training of the GCN \cite{kipf2017} and the GAT \cite{vel2018} models. Serialized weights (\texttt{state\_dict}) and initialization arguments are stored.
    \item \textit{Inference Engine}: The inference engine loads selected models, performs inference on current (original or edited) graph data, generating predictions, and node embeddings.
    \item \textit{Explanation Module}: This module interfaces with \texttt{torch\_geometric.explain.Explainer}. It integrates GNNExplainer for feature/edge masks and extracts GAT attention weights.
    \item \textit{Graph Editing Logic}: Handles front-end requests to modify graph structure, updates graph data in \texttt{dcc.Store}, and triggers re-inference/re-explanation cycles.
\end{itemize}

\subsubsection{Frontend (Dash Application)}
\begin{itemize}
    \item \textit{Layout}: Defines dashboard structure using \texttt{dash.html}, \texttt{dash.dcc}.
    \item \textit{Interactive Components}: \texttt{dcc.Dropdown} (selections), \texttt{dcc.Input}/\texttt{html.Button} (editing), \texttt{dash\_cytoscape.Cytoscape} (graph view), \texttt{dcc.Graph} (Plotly charts).
    \item \textit{State \& Callbacks}: \texttt{dcc.Store} manages state. \texttt{@callback}s link inputs to backend and update views.
\end{itemize}

\subsection{Model Training Details}
\label{subsec:model_training}
\subsubsection{Dataset Preparation} 
We utilize standard Planetoid splits for Cora and CiteSeer. For AmazonPhoto, random node splits 
are generated during offline training. All data sets are pre-processed in \texttt{.pt} files for faster application loading.

\subsubsection{Model Architectures} 
Standard two-layer GCN \cite{kipf2017} (ReLU, dropout) and two-layer GAT \cite{vel2018} (8 heads layer 1, RELU, dropout; single head output layer) are implemented. Their \texttt{forward} methods are adapted for the \texttt{Explainer} framework (returning log-softmax), while separate \texttt{inference} methods provide embeddings and, for GAT, attention weights.

\subsubsection{Training Procedure} 
Models are trained for 200-300 epochs using Adam optimizer \cite{king2015} (LR $\sim$0.005, weight decay $\sim$5e-4) to minimize NLL loss. The best model based on the accuracy of the validation is saved, with early stopping (patience $\sim$20 epochs) to prevent overfitting.

\begin{figure}[htbp]
\centering
\includegraphics[width=1.0\columnwidth]{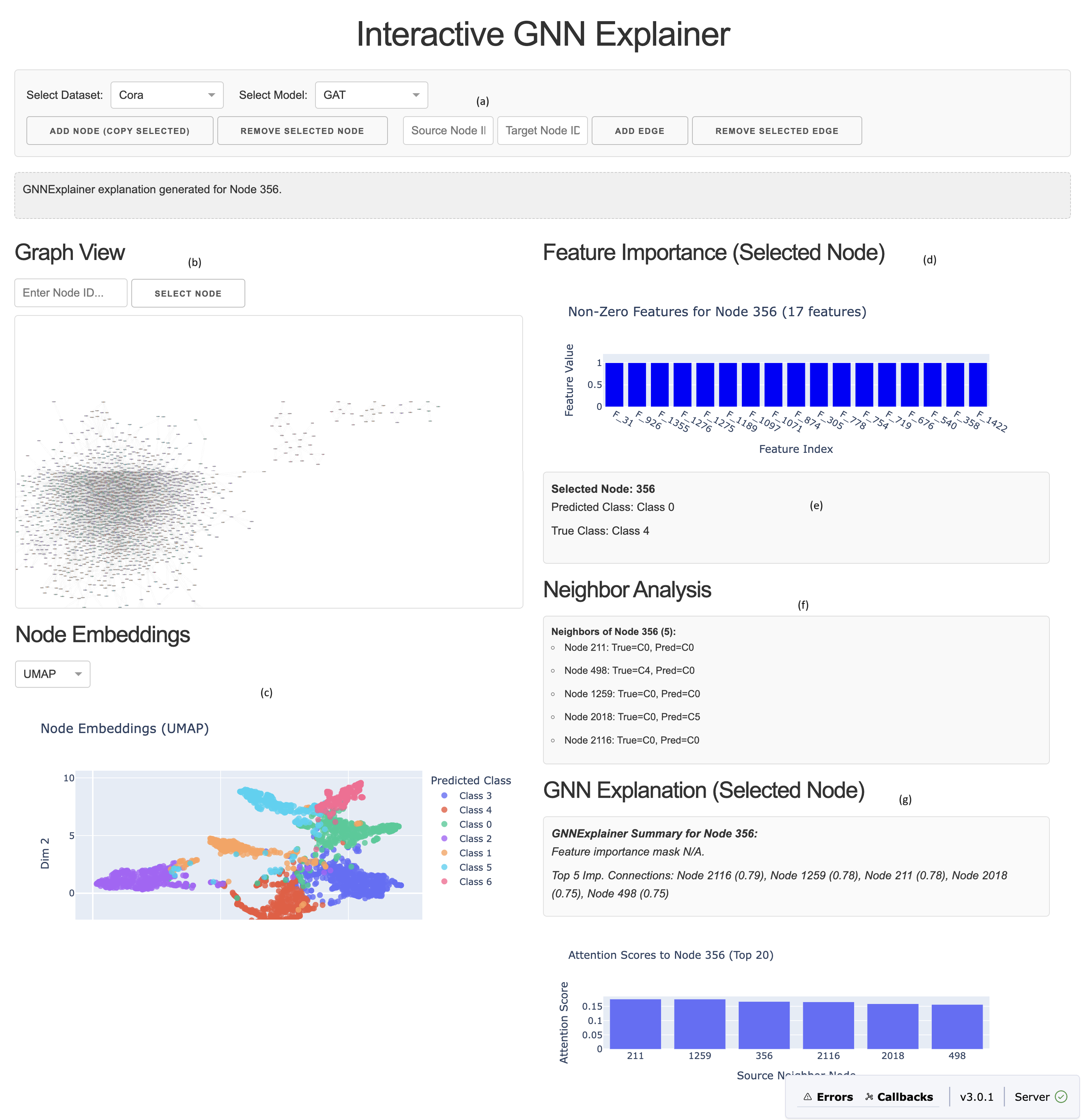} 
\caption{Overview of the InteractiveGNNExplainer dashboard: (a) Control Panel, (b) Graph View, (c) Embeddings View, (d) Feature Importance, (e) Selected Info, (f) Neighbor Analysis, (g) Explanation View.}
\label{fig:dashboard}
\end{figure}
\subsection{Core Interactive Components and Functionality}
Fig.~\ref{fig:dashboard} shows the dashboard of our system. It has the following components:
\begin{enumerate}[label=\alph*]
    \item \textbf{Control Panel (Fig.~\ref{fig:dashboard}a)}: Central hub for data set  (Cora, CiteSeer, AmazonPhoto) and model selection (GCN, GAT). It also incorporates graph editing tools: add node (features from template or default), remove selected node, add edge (by IDs), and remove selected edge.
    \item \textbf{Graph View (Cytoscape) (Fig.~\ref{fig:dashboard}b)}: Primary interactive visualization of the graph structure. Nodes are colored by GNN-predicted class. Selected nodes are highlighted (e.g., larger size, border). Edges identified as important by GNNExplainer are visually distinguished (e.g., green, thicker). Standard interactions like pan, zoom, and node dragging are supported.
    \item \textbf{Node Embeddings View (Plotly) (Fig.~\ref{fig:dashboard}c)}: Displays a 2D projection of the GNN-learned node embeddings. The user can choose various dimension reduction techniques for this view: UMAP \cite{McI2018}, t-SNE \cite{Maa2008}, or PCA. Points are colored by predicted class, mirroring the Graph View. The selected node is emphasized. This view helps assess class separability in latent space and identify outliers or nodes near decision boundaries.
    \item \textbf{Feature Importance View (Plotly) (Fig.~\ref{fig:dashboard}d)}: For a selected node, this presents a bar chart of its input features. It can show raw non-zero feature values or, after GNNExplainer execution, the top-k features identified as most important by the explainer's feature mask, with bars colored by importance.
    \item \textbf{Selected Node/Edge Info Panel (Fig.~\ref{fig:dashboard}e)}: A textual display providing quick details for the selected element: Node ID, its true class label (if available), and its GNN-predicted class label. For edges, it shows source and target node IDs.
    \item \textbf{Neighbor Analysis Panel (Fig.~\ref{fig:dashboard}f)}: When a node is selected, this panel lists its 1-hop neighbors, displaying each neighbor's ID, true class, and GNN-predicted class. This is crucial for understanding the local context and potential influences on the selected node's prediction.
    \item \textbf{Explanation View Panel (Fig.~\ref{fig:dashboard}g)}: Consolidates outputs from explanation algorithms.
        \begin{itemize}
            \item \textit{GNNExplainer Summary}: Textually lists top-k important features and edges (with scores) for the selected node.
            \item \textit{Attention View (GAT only)}: If a GAT model is active, it displays a bar chart of GAT attention scores, showing how the selected node weights its neighbors (or is weighted by them).
        \end{itemize}
    \item \textbf{Interactive Graph Editor}: The core of "what-if" analysis. User edits to the graph (via Control Panel) update the graph data, triggering callbacks for re-running model inference and, if a node is selected, re-generating its explanation. All views dynamically update to reflect these changes, allowing for an iterative exploration of the behavior of the model.
\end{enumerate}

\section{Evaluation: Case studies}
\label{sec:evaluation}
We demonstrate InteractiveGNNExplainer's utility through case studies on Cora and CiteSeer using GCN/GAT models, focusing on how the system facilitates user tasks.

\begin{figure}[htbp]
\centering
\includegraphics[width=0.8\columnwidth]{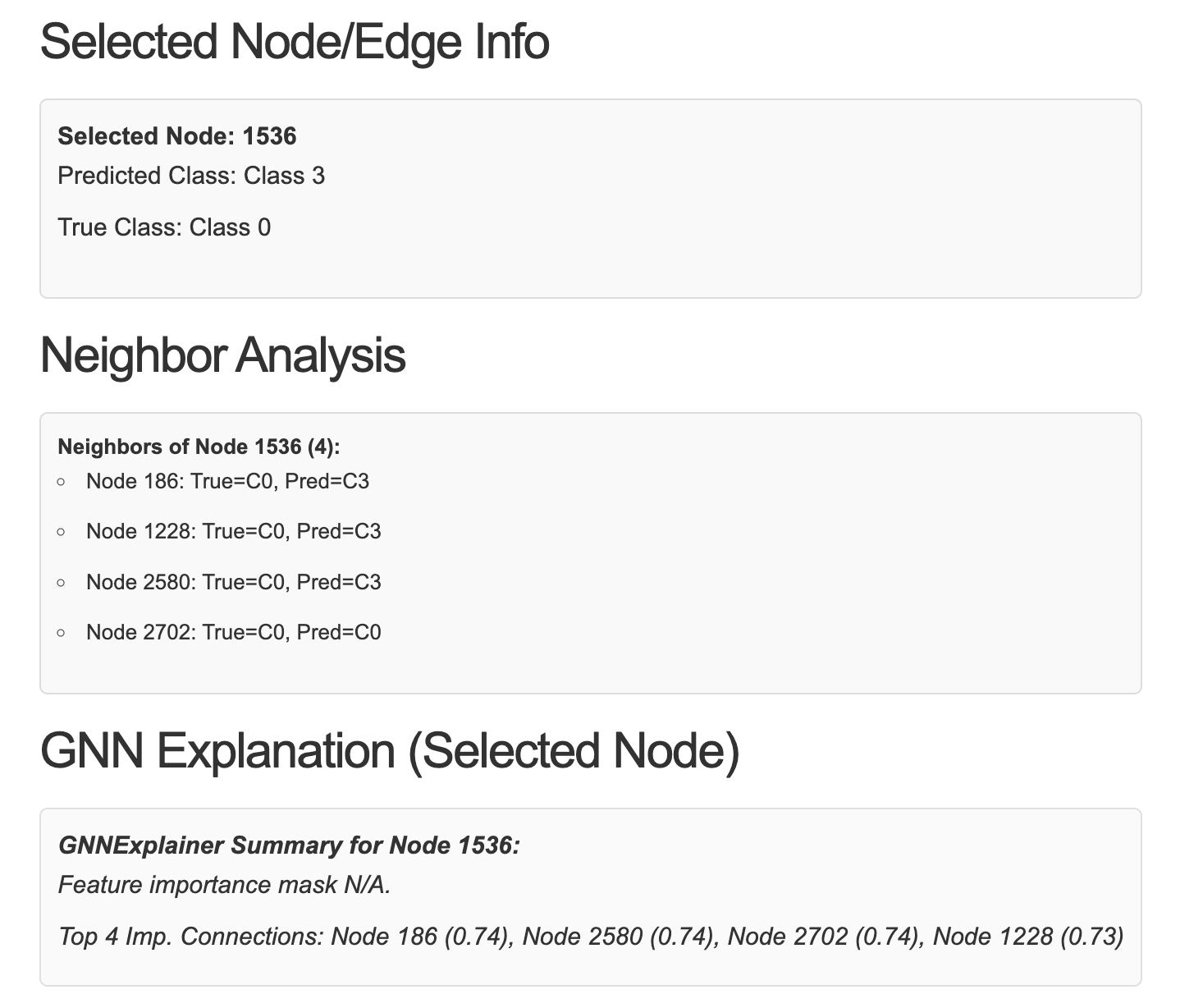}
\caption{Initial analysis of misclassified Node 1536 on Cora. Graph view shows GNNExplainer edge highlights. Neighbor analysis indicates peer misclassifications.}
\label{fig:case1_initial}
\end{figure}
\subsection{Case Study 1: Diagnosing GCN Misclassification on Cora}
\textbf{Objective}: Understand GCN misclassification of node 1536 (True: ``Theory'', Predicted: ``Neural Networks'') and validate explanations via graph editing.

\textbf{Initial Exploration} (Fig.~\ref{fig:case1_initial}): The user selects node 1536. The Neighbor Analysis panel reveals that three of its four neighbors are also mispredicted as ``Neural Networks'', suggesting a strong local influence. GNNExplainer highlights connections to all neighbors as important. The embedding view confirms the proximity of node 1536 to the ``Neural Networks'' cluster.

\textbf{Interactive Probing}: The user hypothesizes that the misclassification is due to the aggregate influence of its misclassified neighbors. To test this, we remove the edge to node 2580, an influential neighbor also misclassified as ``Neural Networks''.

\begin{figure}[htbp]
\centering
\includegraphics[width=0.7\columnwidth]{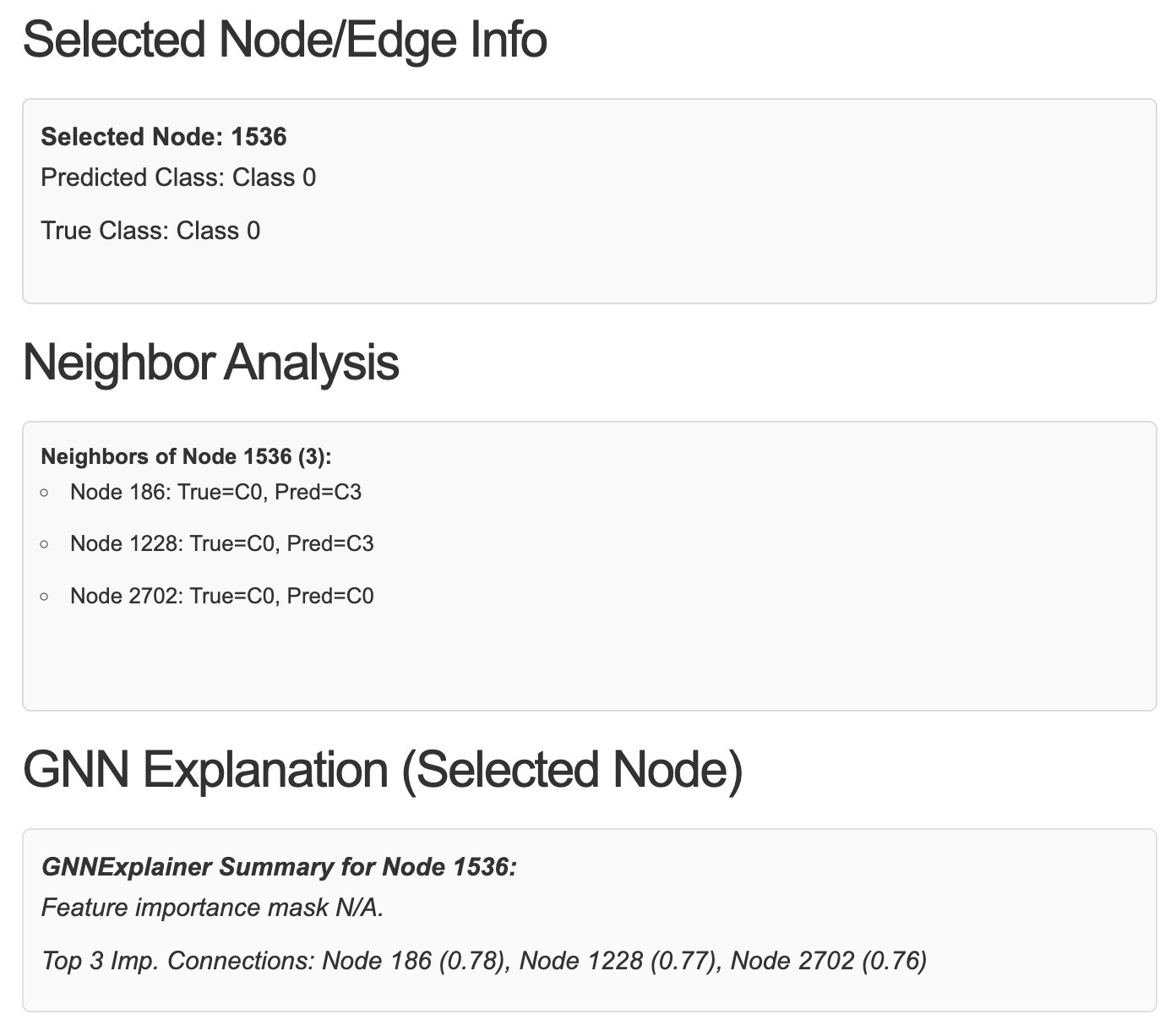} 
\caption{Node 1536 after removing an influential edge. Prediction changes correctly to theory(True) value, reflecting model sensitivity.}
\label{fig:case1_edit}
\end{figure}

\textbf{Observation} (Fig.~\ref{fig:case1_edit}): Upon re-inference, the system shows that node 1536's prediction is corrected to ``Theory''. Its position in the embedding view is also visibly shifted towards the ``Theory'' cluster. A re-run of GNNExplainer would show altered importance for the remaining edges.

\textbf{Insights}: This interactive process allows the user to confirm GNNExplainer's findings regarding edge influence and directly observe the GCN's sensitivity to local neighborhood structure, providing evidence for error propagation.

\subsection{Case Study 2: Comparing GCN and GAT Explanations on CiteSeer}
\textbf{Objective}: Compare how the GCN and GAT models explain the same correct prediction for a selected node in the CiteSeer dataset (e.g., a node correctly predicted to belong to the ``AI'' class).

\textbf{GCN Analysis}: After selecting the GCN model and the target node, GNNExplainer is invoked. The resulting explanation might highlight a somewhat broader set of influential neighboring papers and several general AI-related terms from the node's features, reflecting GCN's uniform aggregation strategy.

\begin{figure}[htbp]
\centering
\includegraphics[width=0.8\columnwidth]{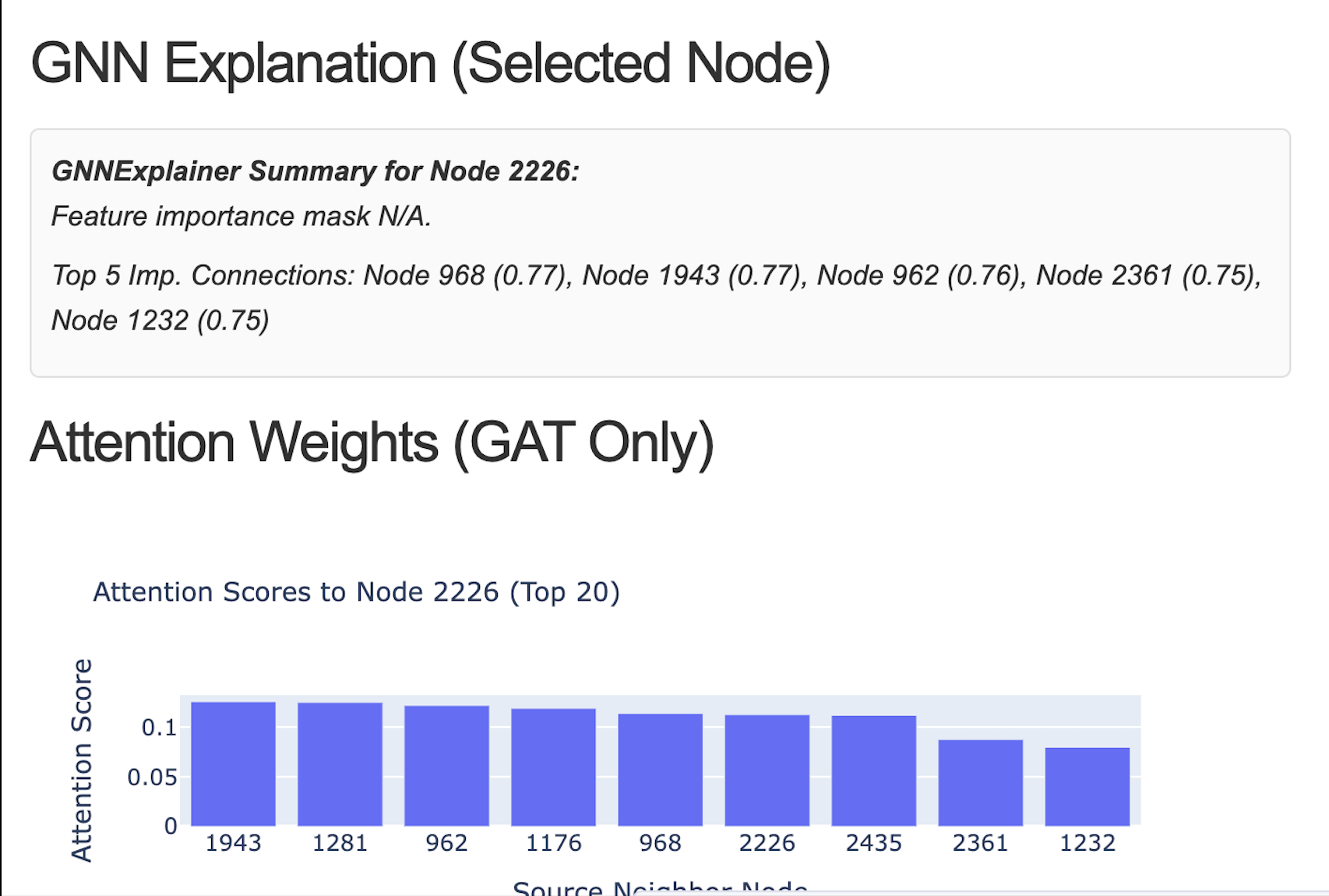} 
\caption{GAT explanation for a node on CiteSeer: GNNExplainer subgraph (top) and GAT attention scores (bottom) illustrate focused importance from the attention mechanism.}
\label{fig:case2_gat}
\end{figure}

\textbf{GAT Analysis} (Fig.~\ref{fig:case2_gat}): The user switches to the GAT model.
\begin{itemize}
    \item GNNExplainer for GAT often yields a sparser explanation, pinpointing fewer, more critical edges and features.
    \item The Attention View directly visualizes GAT's learned attention scores, revealing which specific neighbors the model deemed most important for its prediction about the target node.
\end{itemize}
\textbf{Comparative Insights}: The user can directly observe that GAT's explanations (both from GNNExplainer and its intrinsic attention) tend to be more focused than GCN's. This highlights how GAT's selective attention mechanism differs from GCN's mean aggregation. Users can also assess the consistency between GNNExplainer's post hoc explanation for GAT and GAT's own attention scores. High alignment builds confidence; discrepancies can trigger further investigation into the nuances of both the model and the explainer.

\section{Discussion}
\label{sec:discussion}
InteractiveGNNExplainer advances GNN understanding by addressing our research questions through its integrated, interactive design, offering a significant step toward more transparent graph AI.

\textbf{RQ1 (Enhanced Comprehension)} is achieved by the synergistic effect of coordinated views. Users cross-reference structural (Graph View with GNNExplainer overlays), feature-based (Feature Importance), embedding (Embeddings View), and algorithmic (Explanation View) information. This holistic approach, such as correlating an important edge with an embedding outlier or a specific feature pattern, provides a richer context than isolated explanation methods. For example, a user might observe that a node is misclassified despite its features strongly suggesting the correct class; the neighbor analysis and graph view might then reveal that it is predominantly connected to nodes of the incorrect class, suggesting that structural influence outweighs feature influence. This ability to triangulate evidence from multiple perspectives reduces cognitive load and fosters deeper and more reliable insights into model behavior.

\textbf{RQ2 (Effective Model Probing)} is met through the interactive graph editing feature. This transforms users into active experimenters, enabling validation of hypotheses derived from explanations (e.g., testing if removing an "important" edge, as identified by GNNExplainer, alters the prediction as expected) and direct observation of causal impacts. This ``perturb-observe-explain" cycle promotes a more profound understanding of the sensitivity and robustness of the model to structural changes. Users can explore questions like ``How many {\em good} neighbors does this node need to flip its prediction?" or ``Is the model overly reliant on a single bridge node?" Such explorations go beyond correlation to hint at causation.

\textbf{RQ3 (Facilitating Diagnosis and Comparison)} is supported by tools that contextualize errors and allow for direct comparison of GCN and GAT. For misclassification diagnosis, users can investigate whether errors correlate with neighborhood predictions (error propagation), misleading features, or ambiguous positions in the embedding space. For model comparison, users can observe differences in predictive behavior, the structure of learned embedding spaces, and the nature of explanations (e.g., diffuse for GCN, focused for GAT) on identical instances, thereby understanding how architectural choices impact not just performance but also interpretability and failure modes.

The innovation of InteractiveGNNExplainer lies in its unique synthesis of established explanation methods (post hoc GNNExplainer, intrinsic GAT attention) with direct, interactive graph perturbation, and real-time feedback on both predictions and explanations. While other tools offer components of this, the emphasis on an iterative, human-in-the-loop discovery process where users can actively shape the analysis by modifying the graph and immediately seeing the consequences is a key differentiator. This offers a dynamic, potentially more causal exploration path than static tools or systems with limited interactivity.

\subsection{Limitations} 
Limitations primarily include scalability challenges for very large graphs due to the current necessity of full re-inference and re-explanation after each graph edit, which can introduce latency. The feature assignment mechanism for newly added nodes in the editor is currently simplified (e.g., copying from a template or using zero vectors), which might limit the realism of some ``what-if" scenarios related to novel node types. The scope of integrated explanation algorithms is currently focused on GNNExplainer and GAT attention; incorporating a wider variety could provide more diverse perspectives. Furthermore, the present evaluation is qualitative, based on illustrative case studies. Formal user studies involving GNN practitioners and domain experts are needed for quantitative validation of the framework's effectiveness in improving understanding, debugging efficiency, and overall user trust.

\section{Conclusion and Future Work}
\label{sec:conclusion}
We presented InteractiveGNNExplainer, a visual analytics framework designed to significantly enhance the interpretability and trustworthiness of Graph Neural Networks, specifically for node classification tasks. By uniquely combining multiple coordinated interactive views, established post hoc and intrinsic explanation modalities (GNNExplainer and GAT attention), and, critically, direct graph editing capabilities with real-time feedback, our system facilitates in-depth exploration, diagnosis of misclassifications, and causal probing of GNN prediction mechanisms. Case studies on benchmark datasets like Cora and CiteSeer have demonstrated its utility in fostering a deeper, multifaceted understanding of GNN behavior, which is essential for building trust and developing more robust and accountable graph-based AI systems.

Future work will prioritize several key areas of improvement and expansion to address current limitations and broaden the system's impact. 
\begin{itemize}
\item Improving scalability for larger graphs is a primary concern. This will involve incorporating efficient techniques for graph visualization and computation, developing methods for incremental (rather than full) re-inference and re-explanation after minor graph edits (e.g., by only updating affected subgraphs or using influence functions), and further optimizing backend data handling and processing pipelines. 
\item We aim to enhance the scope of graph editing. This includes implementing more sophisticated feature assignment strategies for newly added nodes (e.g., based on neighborhood aggregation, learned generative models, or allowing users to specify feature vectors) and enabling users to directly edit existing node features to observe their impact on predictions and explanations.
\item Expanding the repertoire of explanation algorithms integrated into the framework is crucial. Incorporating diverse methods such as other prominent perturbation-based techniques (e.g., SubgraphX \cite{yuan2021}), various gradient-based attribution methods adapted for graphs, and automated counterfactual explanation generators (e.g., CF-GNNExplainer \cite{lucic2022}) will provide users with a richer and more comprehensive analytical toolkit, allowing for cross-validation of insights derived from different explainers. 
\item Extending the framework to support model-level explanations - providing insights into the global behavior of the GNN, such as identifying universally important features or common graph patterns (motifs) that consistently drive predictions for entire classes - and to handle other GNN tasks like graph classification or link prediction - would significantly broaden its applicability.
\item Conducting user studies with domain experts (e.g., biologists using GNNs for protein interaction networks) and GNN practitioners is essential. These studies will aim to quantitatively validate the system's effectiveness in improving user understanding of GNNs, facilitating more efficient model debugging, enhancing trust in model predictions, and comparing its usability and utility against existing or baseline explanation interfaces. Such studies will provide valuable feedback for iterative refinement of the system.
\end{itemize}

We believe InteractiveGNNExplainer represents a valuable step towards more transparent and explainable AI. We plan to release the system as an open source tool to encourage its adoption, obtain feedback, and encourage further development and customization by the research community, ultimately contributing to the creation of more understandable, reliable, and trustworthy graph-based analysis.



\bibliographystyle{IEEEtran}
\bibliography{main}

\end{document}